\documentclass{Interspeech}
\usepackage{cite}
\usepackage{amsmath,amssymb,amsfonts}
\usepackage{algorithmic}
\usepackage{graphicx}
\usepackage{textcomp}
\usepackage{xcolor}

\usepackage{multirow}
\usepackage{hyperref}
\usepackage{booktabs}
\usepackage{array}
\usepackage{lineno}
\usepackage{tablefootnote}
\usepackage{tikz}
\usepackage{pgfplots}
\usepackage{threeparttable}
\interspeechcameraready

\title{Adapting Whisper for Parameter-efficient Code-Switching Speech Recognition via Soft Prompt Tuning
\thanks{*Corresponding author: Hao Huang}
\thanks{This work was supported by National Natural Science Foundation of China (62466055) and the National Key R\&D Program of China
(2020AAA0107902).}
}

\author[affiliation={1}]{Hongli}{Yang}
\author[affiliation={2}]{Yizhou}{Peng}
\author[affiliation={1,3,4,*}]{Hao}{Huang}
\author[affiliation={5}]{Sheng}{Li}

\affiliation{School of Computer Science and Technology}{ Xinjiang University}{China}
\affiliation{College of Computing and Data Science}{Nanyang Technological University}{Singapore}
\affiliation{}{Joint International Research Laboratory of Silk Road Multilingual Cognitive Computing}{China}
\affiliation{}{Xinjiang Key Laboratory of Multi-lingual Information Technology}{China}
\affiliation{}{Institute of Science Tokyo}{Japan}

\email{hongli@stu.xju.edu.cn, huanghao@xju.edu.cn}

\keywords{Code-switching speech recognition, Soft prompt tuning, Parameter-efficient, Whisper}

\usepackage{comment}

\begin{document}

\maketitle

\begin{abstract}

Large-scale multilingual ASR models like Whisper excel in high-resource settings but face challenges in low-resource scenarios, such as rare languages and code-switching (CS), due to computational costs and catastrophic forgetting. We explore Soft Prompt Tuning (SPT), a parameter-efficient method to enhance CS ASR while preserving prior knowledge. We evaluate two strategies: 1) full fine-tuning (FFT) of both soft prompts and the entire Whisper model, demonstrating improved cross-lingual capabilities compared to traditional methods, and 2) adhering to SPT’s original design by freezing model parameters and only training soft prompts. Additionally, we introduce SPT4ASR, a combination of different SPT variants. Experiments on the SEAME and ASRU2019 datasets show that deep prompt tuning is the most effective SPT approach, and our SPT4ASR methods achieve  further error reductions in CS ASR, maintaining parameter efficiency similar to LoRA, without degrading performance on existing languages.

% Large-scale multilingual ASR models like Whisper excel in high-resource settings but face challenges in low-resource scenarios, such as rare languages and code-switching (CS), due to computational costs and catastrophic forgetting. We explore Soft Prompt Tuning (SPT), a parameter-efficient method to enhance CS ASR while preserving prior knowledge. We evaluate two strategies: 1) full fine-tuning (FFT) of both soft prompts and the entire Whisper model, demonstrating improved cross-lingual capabilities compared to traditional approaches, and 2) adhering to SPT’s original design by freezing model parameters and only training soft prompts. Additionally, we introduce SPT4ASR, a combination of different SPT variants. Experiments on the SEAME and ASRU2019 datasets show that deep prompt tuning is the most effective SPT approach, and our SPT4ASR methods achieve further error reductions in CS ASR, while maintaining parameter efficiency similar to LoRA and without degrading performance on existing languages.

\end{abstract}

\section{Introduction}

Foundation models (FMs) such as LLaMA\cite{dubey2024llama}, CLIP\cite{RadfordKHRGASAM21}, and Whisper\cite{radford2023robust} have demonstrated strong cross-domain capabilities through large-scale pretraining. Among them, Whisper excels in multilingual automatic speech recognition (ASR), benefiting from massive speech-text data and , showcasing the growing capabilities of publicly accessible speech models in handling complex multilingual tasks.

However, a significant challenge that remains in FMs is how to efficiently adapt to low-resource scenarios without compromising the performance of existing ones\cite{goodfellow2013empirical}. To tackle this issue, previous works in the speech signal processing field have utilized transfer learning techniques, including prompt engineering\cite{wang2024can,peng2023prompting,yang2024prompts}, full fine-tuning (FFT)\cite{yang2024adapting,jain23_interspeech,talafha23_interspeech}, and parameter-efficient tuning (PEFT) methods such as Low-Rank Adaptation  (LoRA)\cite{hu2021lora,song2024lora,xu2024towards}. These approaches leverage the pre-trained weights and knowledge of these FMs, adapting them to new low-resource language environments.

\begin{figure}[tb]
\centering
\includegraphics[width=1.0\linewidth]{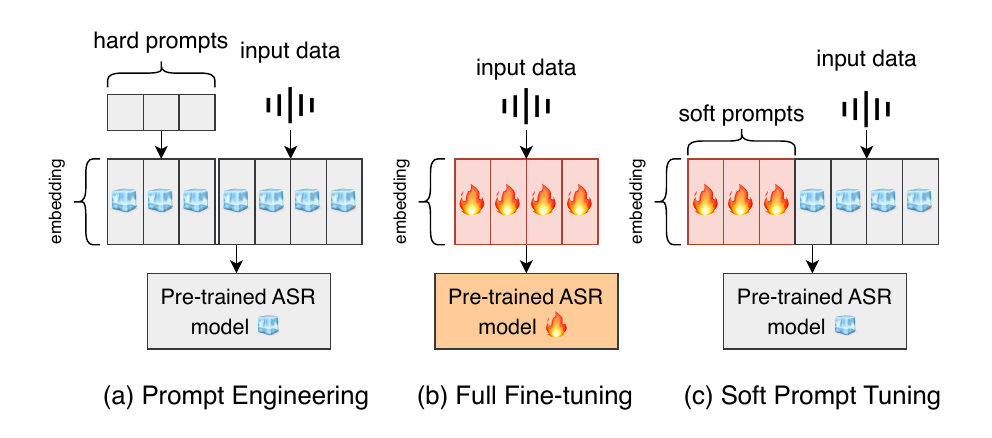}
\caption{Comparison of (a) Prompt Engineering, (b) FFT, and (c) SPT. The fire and ice emojis in the figure represent trainable and frozen parameters, respectively.}
\label{fig:prompt_types}
\end{figure}

Soft Prompt Tuning (SPT) is proposed among PEFT methods for efficiently optimizing a small set of task-specific prompt parameters \cite{lester2021power,chen2023exploring,ts_whisper,qian24_interspeech,ng2024soft}, balancing simplicity, adaptability, and resource conservation similar to LoRA. During fine-tuning with SPT, the parameters of the original model are treated as constants and remain frozen, while the soft prompts are treated as variables and updated during training. Thus, the training process focuses solely on optimizing the soft prompts, enabling efficient and incremental model adaptation. Figure~\ref{fig:prompt_types} illustrates the key differences between prompt engineering, FFT, and SPT.
% Its success has led to variants like Deep Prompt Tuning (DPT)\cite{liu2021p}, Residual Prompt Tuning (ResPT)\cite{razdaibiedina2023residual}, and Language Prompt Tuning (LPT) \cite{li2024enhancing}, each offering unique advantages in model adaptation. 
Recent studies have explored using SPT for cross-lingual transfer, leveraging multilingual language models to transfer knowledge from high-resource to low-resource languages across various NLP tasks \cite{philippy2024soft}. Moreover, some findings suggest that combining soft prompt tuning with partial or full model fine-tuning can further improve performance on downstream tasks \cite{zhao2021discrete,huang2022zero}.

As a low-resource scenario, code-switching (CS) ASR refers to language switching within spontaneous multilingual recordings, including both intra-sentence and inter-sentence cases.  Current approaches to CS ASR include joint optimization methods\cite{abs-2207-04176,PengZXHC22,LiuXGKHK23}, conditional factorization methods\cite{YanZYZDBWWY22,LuHLGQ20,TianYZZ022}, and mixture-of-experts method\cite{YouFSY21}. However, the recent advancements in Large Language Models (LLMs) offer a new approach to these challenges, potentially outperforming task-specific architectures through their inherent understanding of multiple languages and acoustic patterns.

In this paper, we introduce SPT for cross-lingual ASR transfer learning and validate its effectiveness through CS experiments. Our main contributions include: 1) evaluating various SPT-based approaches, including the combination of different variants called SPT4ASR, to enhance CS ASR capability; 2) introducing SPT-based methods that achieves parameter efficiency while mitigating catastrophic forgetting; and 3) to the best of our knowledge, we are the first to conduct a comprehensive comparison of various SPT-based methods for mitigating catastrophic forgetting in multilingual Whisper models.

\begin{figure*}[tb]
\centering
\includegraphics[width=1.0\linewidth]{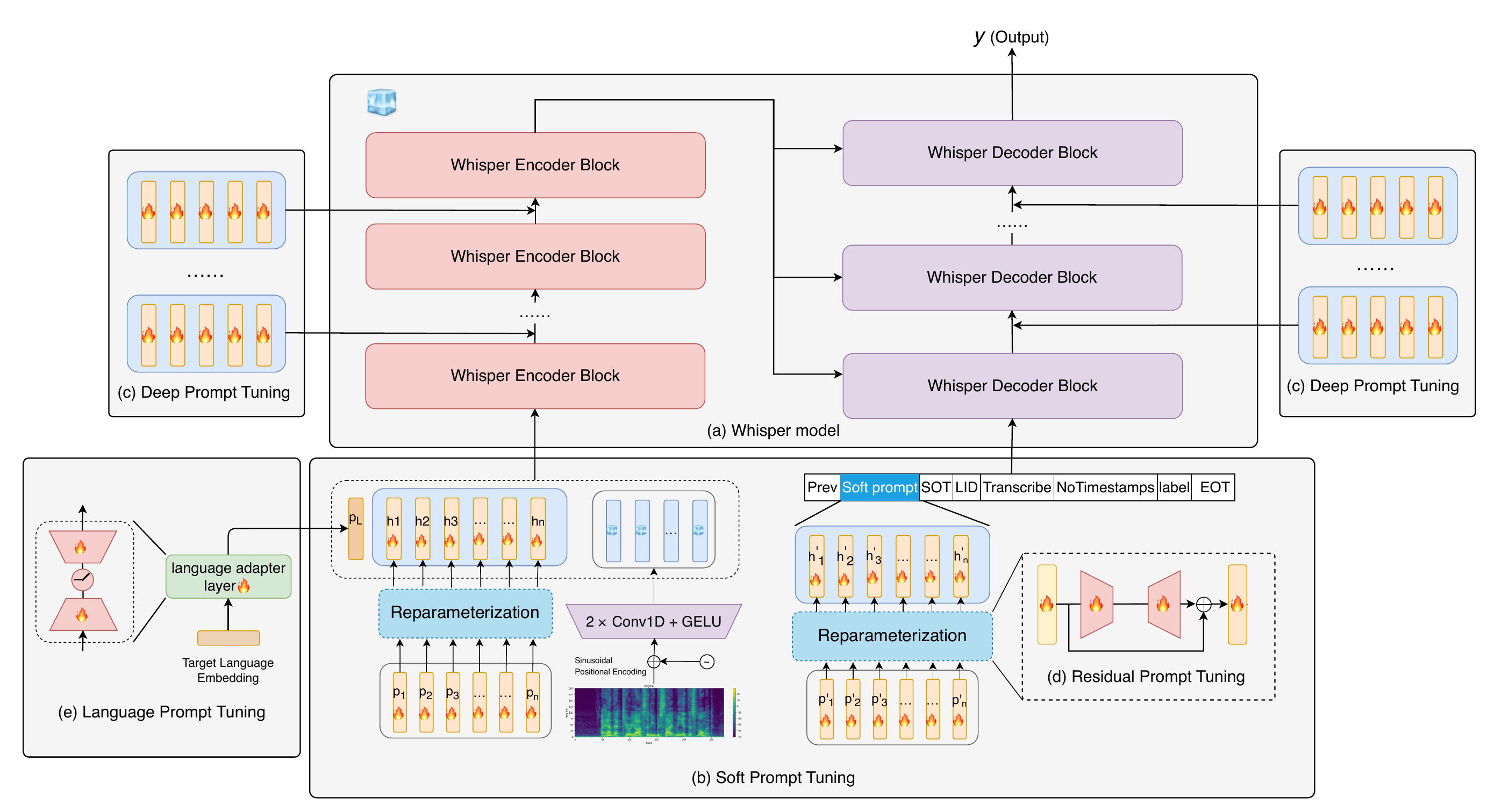}
\caption{Overview of Whisper model and various Prompt Tuning methods: (a) Whisper is a Transformer-based encoder-decoder model for multilingual ASR and translation tasks, where the entire model is frozen in SPT-based experiments. (b) Vanilla Soft Prompt Tuning involves adding prefix embeddings to the model input features. (c) Deep Prompt Tuning (DPT) extends this by adding prefix embeddings to the output of each block within the Whisper model. (d) Residual Prompt Tuning (ResPT) leverages an autoencoder with reparameterization through residual connections to enhance the model. (e) Language Prompt Tuning (LPT) encodes discrete Language IDs (LIDs) into prompt embeddings to improve language discrimination. (f) SPT4ASR is a combination of the aforementioned methods, specifically designed to improve cross-lingual transfer and language expansion in ASR models.}
\label{fig:Soft_prompt_tuning}
\end{figure*}

\section{Methods}

\subsection{Background of Whisper}

Fig.~\ref{fig:Soft_prompt_tuning}(a) shows that the Whisper model uses a Transformer-based encoder-decoder architecture. It excels in multi-lingual speech recognition and translation, handling multiple tasks simultaneously through special task tokens as decoder input prompts. For English ASR without predicting timestamps, the decoder input sequence $g$ is structured as:
\[
\langle\text{Prev}\rangle, \langle|\text{SOT}|\rangle, \langle|\text{EN}|\rangle, \langle|\text{Transcribe}|\rangle, \langle|\text{NoTimestamps}|\rangle,
\]
where \( \langle\text{Prev}\rangle \) optionally represents previous text tokens for context during decoding, and succeeding \( \langle|\cdot|\rangle \) tokens denote for Start of Transcript, Language ID, Transcription task, and No Timestamps specification. 

\subsection{FFT}

% FFT involves fine-tuning all pre-trained model weights for a new task \cite{yang2024adapting}. Given a foundation model $M$ with weights $\theta$ and a new dataset $Y$, the objective is:
% \begin{equation}
% \underset{\boldsymbol{\theta}}{\max } \sum_{(X, y) \in Y} \log P_{\theta} (y \mid X),
% \end{equation}
% where $\theta$ represents all model weights, and $(X, y)$ are input-label pairs in the new dataset $Y$.
FFT is the mainstream method for adapting Whisper to new languages \cite{yang2024adapting,jain23_interspeech,talafha23_interspeech}. In this approach, all parameters of Whisper are optimized using the training data from the new language to enhance its performance in that specific language.
% FFT is the mainstream method for adapting Whisper to new languages \cite{yang2024adapting,jain23_interspeech,talafha23_interspeech}. In this approach, all parameters of Whisper are updated using the new language's training data.
Given a pre-trained Whisper model with parameters $\theta$ and a new dataset $Y$, the FFT objective is:
\begin{equation}
\underset{\boldsymbol{\theta}}{\max } \sum_{(X, y) \in Y} \log P_{\theta} (y \mid X,\theta),
\end{equation}
where $(X, y)$ are speech-label pairs from the dataset $Y$.

\subsection{LoRA}

LoRA is a prominent PEFT technique that efficiently adapts models to new tasks or languages \cite{song2024lora,xu2024towards}. It adjusts only a subset of weights by adding low-rank decomposition matrices to the model, while keeping the original parameters $\theta$ frozen. This approach significantly reduces computational costs and memory requirements compared to FFT, while maintaining comparable effectiveness. The optimization objective for LoRA can be formulated as:
\begin{equation}
\underset{\boldsymbol{\theta}^{\prime}}{\max } \sum_{(X,y)\in Y} \log {P}_{\theta} (y |X,\theta,{\theta}^{\prime} ),
\end{equation}
where $\theta^{\prime}$ represents the subset of trainable or newly added parameters. 
% This formulation demonstrates how PEFT methods focus on optimizing a limited set of parameters, preserving the core knowledge of the pre-trained model.

% To address the limitations of FFT, PEFT methods have been developed. PEFT techniques, such as LoRA\cite{hu2021lora}, Adapters\cite{houlsby2019parameter},  and SPT\cite{lester2021power,song2024lora,ts_whisper}, adjust only a subset of weights $\theta^{\prime}$ or add new parameters, while keeping most original parameters $\theta$ fixed. This approach significantly reduces computational costs and memory requirements compared to FFT, while still effectively adapting the model to new tasks. PEFT methods offer a more efficient way to leverage the model's pre-trained knowledge for new task adaptation.

% As an example of PEFT methods, the optimization objective for LoRA can be formulated as:
% \begin{equation}
% \underset{\boldsymbol{\theta}^{\prime}}{\max } \sum_{(X,y)\in Y} \log {P}_{\theta} (y |X,\theta,{\theta}^{\prime} ),
% \end{equation}
% where $\theta^{\prime}$ represents the subset of trainable or newly added parameters. This formulation demonstrates how PEFT methods focus on optimizing a limited set of parameters, allowing for efficient model adaptation while preserving the core knowledge of the pre-trained model.

\subsection{SPT}

% Among PEFT methods, SPT stands out for its simplicity and efficiency. SPT introduces learnable embeddings (soft prompts) that are concatenated to the model's existing input \( X \) as a prefix or suffix, while keeping all other model parameters \( \theta \) frozen. 
% This approach leverages pre-existing knowledge in the PLMs while reducing the reliance on extensive task-specific fine-tuning.
SPT adapts Whisper by introducing learnable embeddings \( P \), namely soft prompts, which are concatenated to the model's inputs while freezing Whisper model parameters $\theta$. Specifically, SPT optimizes a trainable embedding matrix \( P \in \mathbb{R}^{n \times e} \), where \( n \) represents the number of soft prompts and \( e \) is the dimension of each embedding. The optimization problem is formulated as follows:

\begin{equation}
\underset{\boldsymbol{P}}{\max } \sum_{(X,y)\in Y} \log {P}_{\theta} (y |[P,X],\theta) .
\end{equation}

Applying SPT to Whisper involves designing separate soft prompt sequences for both the encoder and decoder. Fig.~\ref{fig:Soft_prompt_tuning} (b) illustrates this process. Given a sequence of acoustic features $ X \in \mathbb{R}^{l \times e}$ and a decoder embedding matrix $E \in \mathbb{R}^{|V| \times e}$, we determine the optimal soft prompts $P \in \mathbb{R}^{n \times e}$ for the encoder input and $P^{\prime} \in \mathbb{R}^{n \times e}$ for the decoder input. Here, $l$ is the input speech length, $|V|$ is the vocabulary size, and $n$ is the length of soft prompts. 

We experimented with different SPT implementations by adding soft prompts to the encoder, decoder, or both. For encoder adaptation, the model concatenates $[P, X] \in \mathbb{R}^{(n+l) \times e}$, where $P$ is the soft prompt matrix of length $n$, $X$ is the input features of length $l$, and $e$ is the embedding dimension. This allows the encoder to process both the original input features and the soft prompts together. For decoder adaptation, soft prompts $P^{\prime} \in \mathbb{R}^{n \times e}$ are inserted into the $\langle\text{Prev}\rangle$ token of the decoder input sequence $g$, which includes other special tokens such as start-of-sequence, language ID, and task tokens. The resulting input is $[P^{\prime}, \text{embed}(g)] \in \mathbb{R}^{(n+4) \times e}$, where $n+4$ represents the total number of tokens in the modified sequence. This modified input provides additional context to the decoder, guiding the decoding process effectively.

\subsubsection{Deep Prompt Tuning}

In the above vanilla SPT approach, where soft prompts are only concatenated with input features before entering the encoder and decoder, we note some limitations:  the prompt length is constrained by computational costs, limiting instruction capacity, and the static prompts are too distant from the output layers to directly influence final predictions. 

Fig.~\ref{fig:Soft_prompt_tuning} (c) shows that we introduce Deep Prompt Tuning (DPT) \cite{liu2021p}, a novel approach that inserts soft prompts at various intermediate blocks throughout the Whisper model. By strategically placing prompts across different levels of the network, we enable a more direct and nuanced influence on the model's predictions without altering the original parameters. 

% To overcome these limitations, we introduce Deep Prompt Tuning, a novel approach that inserts soft prompts at various intermediate blocks throughout the Whisper model \cite{liu2021p}. As depicted in Fig. \ref{fig:Soft_prompt_tuning} (c), this method not only preserves the model's pre-trained capabilities but also significantly enhances its adaptability to new tasks. By strategically placing prompts across different levels of the network, we enable a more direct and nuanced influence on the model's predictions without altering the original parameters. 

% Unlike the vanilla approach, where prompts are confined to the input layer, Deep Prompt Tuning leverages the power of continuous prompts at each block of the pre-trained model. This distribution of prompts throughout the network architecture allows for a more profound impact on model predictions, effectively addressing the limitations of static, input-layer-only prompts.

\subsubsection{Residual Prompt Tuning}

To overcome limitations of traditional SPT, such as slow convergence and reliance on pre-training prompts, we introduce Residual Prompt Tuning (ResPT) \cite{razdaibiedina2023residual}, inspired by reparameterization. As illustrated in Fig.~\ref{fig:Soft_prompt_tuning} (d), ResPT employs a shared multi-layer perceptron (sharedMLP) for reparameterization, enabling efficient and consistent prompt generation. The sharedMLP reparameterizes all prompts, improving training efficiency and robustness through parameter sharing.

ResPT's key advantage is refining soft prompts quickly without requiring extensive pre-training. The use of residual connections allows the model to build on previous prompt information, while the sharedMLP ensures stable learning. Mathematically, the reparameterization is given by:

\begin{equation}
\mathbf{P}^{\prime} = \operatorname{MLP}(\mathbf{P}) + \mathbf{P},
\end{equation}
where \(\mathbf{P}\) represents the input soft prompts, \(\operatorname{MLP}(\mathbf{P})\) is the output of the shared MLP, and \(\mathbf{P}^{\prime}\) is the refined soft prompt.

% To further accelerate convergence and enhance training efficiency, we introduce Residual Prompt Tuning, inspired by the work of reparameterization\cite{razdaibiedina2023residual}. This method addresses the limitations of traditional soft prompts, particularly their slow convergence and the need for pre-training. As illustrated in Fig. \ref{fig:Soft_prompt_tuning} (d), our approach utilizes a shared multi-layer perceptron (sharedMLP) as our reparameterization scheme to optimize performance and ensure consistency across the model. 
% This unified ResMLP reparameterizes all prompts, simplifying the training process and enhancing robustness through parameter sharing across different blocks. 
% Moreover, this approach improves the model's adaptability to new tasks and languages, increasing the effectiveness of transfer learning. 
% The reparameterization process can be formulated as:
% \begin{equation}
% \mathbf{P}^{\prime}=\operatorname{MLP}(\mathbf{P})+\mathbf{P},
% \end{equation}
% where $\mathbf{P}$ represents the input Soft prompts, $\operatorname{MLP}(\mathbf{P})$ denotes the output of the shared MLP, and $\mathbf{P}^{\prime}$ is the refined Soft prompts enhanced by residual connections.

\subsubsection{Language Prompt Tuning} 

Building on Whisper's language identification capabilities and prior work on language token understanding \cite{peng2023prompting,yang2024adapting,yang2024prompts}, we introduce Language Prompt Tuning (LPT) \cite{li2024enhancing} to enhance code-switching ASR performance. Our approach combines this with a Language Encoder, as an adapter module designed to improve language differentiation in code-switching environments. The Language Encoder is a lightweight, task-specific component that integrates context-specific cues for each language, allowing the model to dynamically adapt to language transitions more efficiently and accurately. 

% As illustrated in Fig. \ref{fig:Soft_prompt_tuning} (e), these language-specific embeddings are concatenated with acoustic features and fed into the end-to-end multilingual speech recognition model.

% Building on Whisper's language identification capabilities and previous research \cite{peng2023prompting,yang2024adapting,yang2024prompts}, we introduce Language Prompt Tuning to enhance code-switching ASR performance. This method encodes discrete Language IDs (LIDs) into prompt embeddings, functioning similarly to task IDs, to improve language discrimination. As shown in Fig. \ref{fig:Soft_prompt_tuning} (e), these language-specific embeddings are concatenated with acoustic features before being fed into the model \cite{li2024enhancing}.
In our implementation of CS ASR, we integrate two LIDs to fully exploit Whisper's capability to interpret language tokens. Specifically, we utilize Whisper's pre-trained language embeddings as prompts, guiding the model to more accurately identify distinct languages in mixed-language speech contexts. As shown in Fig.~\ref{fig:Soft_prompt_tuning} (e), these language-specific embeddings are concatenated with acoustic features before being input into the Whisper model. This method not only improves language differentiation in new CS scenarios but also maintains overall ASR performance across multilingual environments.

\begin{table}[ht]
\centering
\scriptsize
\caption{Corpus Details with SEAME and ASRU2019}
\label{tab:corpus_details}
\setlength{\tabcolsep}{5.0mm}
\renewcommand{\arraystretch}{1.2}
\begin{tabular}{l|l|c|c}
\hline
Corpus & Subset & Duration(Hrs) & Man/Eng \\ 
\hline
\hline
\multirow{3}{*}{SEAME} & Train & 93.6 & 0.83/0.17 \\
                       & DevMan & 7.5 & 0.91/0.09 \\
                       & DevSge & 3.9 & 0.55/0.45 \\ \hline
\multirow{2}{*}{ASRU2019}  & Train & 193.0 & 0.97/0.03 \\
                       % & Dev1 & 20.4 & 0.97/0.03 \\
                       % & Dev2 & 21.3 & 0.98/0.02 \\
                       & Test & 20.6 & 0.98/0.02 \\ \hline
                       \hline
\end{tabular}
\end{table}

\begin{table*}[tb]
\centering
\scriptsize
\caption{Speech recognition results on SEAME dataset with different prompt lengths and positions (MER \%)}
\label{tab:prompt_length_position_selection}
\setlength{\tabcolsep}{3.00mm}
\renewcommand\arraystretch{1.1}
\begin{tabular}{l|c|c|c|c|c|c|c|c|c|c}
\hline
% \toprule
% \multirow{2}{*}{Method} & \multicolumn{2}{c|}{16} & \multicolumn{2}{c|}{32} & \multicolumn{2}{c|}{64} & \multicolumn{2}{c|}{128} & \multicolumn{2}{c}{256} \\
\multirow{2}{*}{Prompt Positions} & \multicolumn{2}{c|}{Prompt Length of 16} & \multicolumn{2}{c|}{Prompt Length of 32} & \multicolumn{2}{c|}{Prompt Length of 64} & \multicolumn{2}{c|}{Prompt Length of 128} & \multicolumn{2}{c}{Prompt Length of 256} \\
% \cmidrule(lr){2-3} \cmidrule(lr){4-5} \cmidrule(lr){6-7} \cmidrule(lr){8-9} \cmidrule(lr){10-11}% 这行添加横线
\cline{2-11}  
 & devman & devsge & devman & devsge & devman & devsge & devman & devsge & devman & devsge \\
% \midrule
% \midrule
\hline
\hline
Vanilla Soft Prompt (Encoder) & 27.59 & 32.11 & 27.64 & 35.49 & 32.12 & 37.29 & 24.62 & 34.81 & 24.98 & 34.71 \\
Vanilla Soft Prompt (Decoder) & 32.15 & 45.27 & 30.23 & 42.86 & 27.95 & 40.56 & 25.44 & 39.21 & N/A$^{{\spadesuit}}$ & N/A$^{{\spadesuit}}$ \\
Vanilla Soft Prompt (Entire) & 31.25 & 36.75 & 26.39 & 34.47 & 23.21 &  28.92 & \textbf{21.95} & \textbf{27.39} & 22.31 & 30.81 \\
% \hspace{0.6em} + Deep Prompting & & & & & & & & & & & \\
% \hspace{1.2em} ++ sharedMLP & 19.41 & 26.29 & 17.60 & 25.18 & 17.13 & 29.91 & 16.26 & \textbf{22.99} & \textbf{15.84} & 25.50 & \\
% \hspace{1.2em} ++ sepMLP & 18.73 & 26.19 & 17.95 & 25.15 & 17.23 & 23.82 & \textbf{16.93} & 23.26 & 17.03 & \textbf{22.93} & \\
% \bottomrule
\hline
\hline
\end{tabular}

\begin{tablenotes}    
    \scriptsize               
    \item \textit{Note}: ${\spadesuit}$ indicates the decoder doesn’t support 256 token length due to context limitations. The \textit{Entire} Position uses 256 tokens for the encoder and 128 for the decoder.
\end{tablenotes}

\end{table*}

\section{Experiments}
\label{sec:exp}

\subsection{Dataset and Evaluation Metric}
\label{sec:Dataset}

% Following the approach used in Whisper’s FFT\cite{yang2024adapting}, we optimized soft prompts using the SEAME corpus, a conversational Mandarin-English dataset of approximately 100 hours from Southeast Asia, i.e., Malaysia and Singapore, and the ASRU2019 dataset, a 200-hour Mandarin-English CS dataset from Mainland China, released by Datatang for a Mandarin-English ASR challenge. Both the SEAME and ASRU2019 datasets consist of intra-sentential code-switching speech data and detailed information of the datasets of these three languages are presented in Table \ref{tab:corpus_details}. We adopt the Mix Error Rate (MER) for evaluating CS ASR performance, calculating errors at the character level for Mandarin and at the word level for English. To assess alleviating catastrophic forgetting, we used the Aishell1 dataset for Chinese ASR, the LibriSpeech test set for English ASR, and the Korean language evaluation data from the SPREDS-U1 dataset\footnote{\scriptsize \url{https://ast-astrec.nict.go.jp/en/release/SPREDS-U1}} for Korean ASR.

Following the approach used in Whisper’s FFT\cite{yang2024adapting}, we optimized soft prompts using the SEAME corpus, a conversational Mandarin-English dataset of approximately 100 hours from Southeast Asia, i.e., Malaysia and Singapore, and the ASRU2019 dataset, a 200-hour Mandarin-English CS dataset from Mainland China, released by Datatang for a Mandarin-English ASR challenge. Both the SEAME and ASRU2019 datasets consist of intra-sentential code-switching speech data and detailed information of the datasets of these three languages are presented in Table \ref{tab:corpus_details}. We adopt the Mix Error Rate (MER) for evaluating CS ASR performance, calculating errors at the character level for Mandarin and at the word level for English. To assess alleviating catastrophic forgetting, we used the Aishell1 dataset for Chinese ASR, the LibriSpeech test set for English ASR, and the Korean language evaluation data from the SPREDS-U1 dataset for Korean ASR.

% Following the approach used in Whisper’s FFT\cite{yang2024adapting}, we optimized soft prompts using the SEAME corpus\cite{lyu2015mandarin}, a conversational Mandarin-English dataset of approximately 100 hours from Southeast Asia, i.e., Malaysia and Singapore, and the ASRU2019 dataset\cite{shi2020asru}, a 200-hour Mandarin-English CS dataset from Mainland China, released by Datatang for a Mandarin-English ASR challenge. Both the SEAME and ASRU2019 datasets consist of intra-sentential code-switching speech data and detailed information of the datasets of these three languages are presented in Table \ref{tab:corpus_details}. We adopt the Mix Error Rate (MER) for evaluating CS ASR performance, calculating errors at the character level for Mandarin and at the word level for English. To assess alleviating catastrophic forgetting, we used the Aishell1 dataset for Chinese ASR, the LibriSpeech test set for English ASR, and the Korean language evaluation data from the SPREDS-U1 dataset for Korean ASR.

\subsection{Implementation Details}
\label{sec:Implementation}

% To verify the effectiveness of SPT, we conducted experiments on Whisper’s small and medium models. 
% The small model features 6 encoder and decoder blocks with hidden feature dimensions of 768. The medium model has 12 blocks with hidden feature dimensions of 1024. 
% Both models use 80-dimensional Mel frequency bins as input features, computed on 25ms windows with a 10ms stride.
% We trained the models over 10 epochs using a batch size of 8. 
We conducted experiments on Whisper’s small and medium models, training for 10 epochs with a batch size of 8. 
The AdamW optimizer was used, starting with a learning rate of 1e-3 for LoRA and SPT, and 1e-6 for FFT. The training was conducted on a single NVIDIA RTX3090 GPU with 24GB VRAM. 
For decoding, we selected the best model from all epochs and performed greedy search. 
% The ground truth labels and model outputs were normalized using Whisper’s Normalizer that converts all characters to lowercase and removes punctuation.
% , following Whisper’s processing method.

\section{Results}
\label{sec:result}

\subsection{Prompt Length and Position Selection}
\label{ssec:length}

The performance of SPT is highly sensitive to prompt length. We experimented with Vanilla SPT tuning using lengths \{16, 32, 64, 128, 256\} on the SEAME dataset with the Whisper small model. As shown in Table~\ref{tab:prompt_length_position_selection}, performance improves with increasing length up to 128, the maximum length the decoder can handle due to context length limitations. Notably, tuning the entire model yields the best results. Therefore, in subsequent experiments, we used a prompt length of 128 and applied tuning to the entire model.

\subsection{Main Results}
\label{ssec:subhead}

\begin{table}[tb]
\centering
\scriptsize
\caption{ASR performance comparison results of SPT and other methods on SEAME and ASRU2019 datasets (MER \%)}
\label{tab:Comparison}
\setlength{\tabcolsep}{1.50mm}
\renewcommand\arraystretch{1.2}
\begin{tabular}{l|p{2.45cm}|c|c|c|c}
\hline
\multirow{2}{*}{ID} & \centering\arraybackslash \multirow{2}{*}{Method} & \multicolumn{2}{c|}{SEAME} & \multicolumn{1}{c|}{ASRU} & \multirow{2}{*}{\#Params (M)}  \\
\cline{3-5}
 & & devman & devsge & test &  \\
\hline
\hline
B1 & Whisper Small & 95.59  & 75.66  & 29.74 & / \\
B2 & Whisper Medium &  76.49 &  65.36 &  23.21 & / \\
B3 & Whisper Large-v3 & 61.18 & 60.14 & 18.41 & / \\
B4 & ESPnet Conformer & 16.60 & 23.30 & 13.20 & 47.27M \\
\hline
& \multicolumn{5}{c}{\textbf{Whisper-small Tuning Results}} \\ 
\hline
S1 & FFT & 13.37 & 19.42 & 12.92 & 240.58M \\
S2 & LoRA & 13.96 & 20.59 & 13.00 & 1.85M \\
S3 & Vanilla SPT & 21.95 & 27.39 & 15.60 & 0.20M \\
S4 & \hspace{0.6em}+ DPT & 16.57 & 23.99 & 13.52 & 1.39M \\
S5 & \hspace{0.6em}+ ResPT & 20.23 & 26.43 & 15.06 & 0.79M \\
S6 & \hspace{0.6em}+ LPT & 18.24 & 26.68 & 14.68 & 0.99M \\
% S7 & \hspace{0.6em}++ ResPT & 18.29 & 25.23 & 14.37 & 1.58M \\
S7 & SPT4ASR & 15.48 & 21.98 & 13.12 & 3.74M \\
S8 & Whole model SPT & \textbf{13.07} & \textbf{18.65} & \textbf{12.84} & 240.69M \\
\hline
& \multicolumn{5}{c}{\textbf{Whisper-medium Tuning Results}} \\ 
\hline
M1 & FFT & 11.44 & 17.55 & 10.18 & 762.32M \\
M2 & LoRA & 11.68 & 16.99 & 10.35 & 4.94M \\
M3 & SPT4ASR & 13.12 & 18.10 & 11.30 & 7.48M \\
M4 & Whole model SPT & \textbf{11.17} & \textbf{16.61} & \textbf{10.18} & 762.59M \\
\hline
\hline
\end{tabular}

\begin{tablenotes}    
    \scriptsize               
    \item  \textit{Note}: \#Params refers to the number of trainable parameters.
\end{tablenotes}
\end{table}

% in terms of performance on the SEAME dataset and parameter efficiency.

% Table~\ref{tab:Comparison} presents a comprehensive comparison of SPT variants, FFT, and LoRA methods on the SEAME and ASRU2019 datasets for both Whisper-small and Whisper-medium models. All fine-tuning approaches significantly outperform the zero-shot performance of baseline Whisper models (B1-B3).

Table~\ref{tab:Comparison} presents a comparison of tuning methods, including SPT variants, FFT, and LoRA, on the SEAME and ASRU2019 datasets for Whisper-small and Whisper-medium models, along with zero-shot results from three different Whisper model versions (B1-B3) and the ESPnet Conformer results (B4).

\subsubsection{Comparison of different SPT variants}

As shown in Table~\ref{tab:Comparison}, among the SPT variants for Whisper-small, DPT (S4) achieves the most significant improvement, significantly outperforming the zero-shot performance of baseline Whisper models (B1-B3), and performing similarly to the ESPnet Conformer results (B4). This enhancement can be attributed to its ability to introduce learnable parameters at multiple layers of the model, thus enabling more granular adaptation across the network hierarchy. ResPT (S5) and LPT (S6) also demonstrate substantial enhancements with fewer parameters. Notably, SPT4ASR, a combination of multiple SPT variants (S7), achieves the best performance among SPT-based methods, with only about 3.74M parameters, highlighting the potential and effectiveness of integrating different SPT approaches.

% Among the SPT variants for Whisper-small, the Deep Prompt method (S4) achieves significant improvement, reducing the MER by approximately 83\% on the SEAME devman set compared to the zero-shot baseline(B1). This enhancement can be attributed to its ability to introduce learnable parameters at multiple layers of the model, thus enabling more granular adaptation across the network hierarchy. The Residual Prompt (S5) and Language Prompt (S6) methods also demonstrate significant enhancements while using fewer parameters, with the Residual Prompt being particularly efficient at less than 1M parameters. Notably, the combination of multiple SPT variants (S8) achieves the best performance among SPT-based methods, with only about 3.7M parameters, highlighting the potential and effectiveness of integrating different SPT approaches.

% To visualize the trade-off between parameter efficiency and performance, Fig.~\ref{fig: tradeoff} illustrates the diverse range of tuning methods on SEAME devman set, from highly parameter-efficient to high-performing options. Notably, Whole Model SPT achieves the best overall performance, slightly outperforming FFT. This visualization highlights SPT methods' flexibility in balancing performance and efficiency across various computational requirements similar to LoRA.

\subsubsection{Comparison of SPT with FFT and LoRA}
As shown in Table~\ref{tab:Comparison}, across both Whisper-small and Whisper-medium models, the SPT4ASR methods (S7 and M3) perform slightly worse than FFT while utilizing considerably fewer parameters. 
However, it is worth acknowledging that, in terms of both parameter efficiency and performance, LoRA (S2 and M2) still outperforms SPT4ASR (S7 and M3) , whether in the small or medium model. 
% This highlights the scalability and efficiency of SPT approaches across different model sizes. Notably, these combined SPT variants achieve competitive performance with only about 3.74M parameters for Whisper-small, highlighting the potential of integrating different SPT approaches in a highly efficient manner. 
Surprisingly, we found that when applying Vanilla soft prompts to Whisper-small and Whisper-medium models, integrating soft prompts with whole model full tuning resulted in significantly better performance (S8 and M4) compared to standard FFT methods (S1 and M1). For instance, the Whisper-medium model with both whole model FFT and SPT(M4) achieved a devman MER of 11.17\% and a devsge MER of 16.61\%, outperforming the Medium FFT approach(M1) with only a 0.27M increase in parameters.

\subsection{Mitigation of Catastrophic Forgetting}
\label{ssec:Forgetting}

% Method? Should be prompt position right?
\begin{table}[tb]
\centering
\scriptsize
\caption{SPT results on AISHELL-1 (CER\%), LibriSpeech (WER\%) and Korean (WER\%) from the SPREDS-U1 datasets in mitigating catastrophic forgetting}
\label{tab:Forgetting}
\setlength{\tabcolsep}{1.2mm}
\renewcommand\arraystretch{1.2}
\begin{tabular}{l|p{2.0cm}|c|c|c|c|c} % 修改列格式
% \toprule
\hline
\multirow{2}{*}{ID} & \centering \multirow{2}{*}{Method} & \multicolumn{2}{c|}{Aishell} & \multicolumn{2}{c|}{Librispeech} & \multicolumn{1}{c}{SPREDS} \\
% \cmidrule(lr){3-4} \cmidrule(lr){5-6} % 修改横线的位置
\cline{3-7}
 & & dev & test & test clean & test other & 04kor \\
% \midrule
% \midrule
\hline
\hline
C1 & Whisper Small & 12.72 & 13.91 & 3.40 & 7.60 & 18.90 \\
% C2 & Whisper Small + FFT & 27.41  & 29.32  & 18.17  & 30.21  \\
% C3 & Whisper Small + LoRA & 12.94 & 14.07 & 4.09 & 8.62 \\
% C4 & Whisper Small + SPT Variants & 12.74 & 13.97 & 4.30 & 8.81 \\
C2 & Whisper Medium & 9.38 & 9.54 & 2.90 & 5.90 & 15.96 \\
C3 & ~~ + FFT & 15.82 & 17.16 & 13.15 & 22.17 & 436 \\
C4 & ~~ + LoRA & 9.37 & 9.73 & 3.51 & 6.97 & 16.80 \\
C5 & ~~ + SPT4AS & 9.44 & 9.70 & 3.50 & 7.46 & 17.03 \\
% \bottomrule
\hline
\hline
\end{tabular}
\end{table}

To assess the effectiveness of SPT in alleviating catastrophic forgetting, we compared it with FFT and LoRA methods on monolingual ASR tasks using Whisper Medium models. Table~\ref{tab:Forgetting} shows that while FFT-based models (C3) exhibited significant performance degradation, both SPT4ASR (C5) and LoRA (C4) maintained performance close to the baseline Whisper Medium model (C2). This trend is consistent across Whisper Small and Medium models. This preservation of performance was consistent in both monolingual ASR experiments, suggesting that SPT and LoRA effectively mitigate catastrophic forgetting. The efficacy of SPT and LoRA can be attributed to their plug-and-play nature, allowing for model adaptation without parameter expansion.

% To assess the effectiveness of SPT in alleviating catastrophic forgetting, we compared it with FFT and LoRA methods. We evaluated the performance of our fine-tuned Whisper Medium models on Chinese and English ASR tasks, focusing on their ability to retain original language capabilities after fine-tunning.

% Table \ref{tab:Forgetting} shows the results that FFT-based model (C6) exhibited significant performance degradation, particularly in English ASR tasks. Conversely, both SPT variants (C8) and LoRA (C7) demonstrated robust retention of original language capabilities, with performance close to the baseline Whisper Medium model (C5). The trend of Whisper Small and Medium models is consistent. This preservation of performance was consistent in both Chinese and English ASR experiments, suggesting that SPT and LoRA effectively mitigate catastrophic forgetting. The efficacy of SPT and LoRA can be attributed to their plug-and-play nature, allowing for model adaptation without parameter expansion. This characteristic enhances their utility in developing versatile, multilingual ASR systems while maintaining computational efficiency.

% Overall, these findings suggest that SPT-based methods, especially when combined with variants, offer a promising approach to fine-tuning models in a way that mitigates catastrophic forgetting. This capability is crucial for maintaining robust performance across multiple languages, especially in the context of multilingual ASR systems.

\vspace{-1mm}
\section{Conclusion}
\label{sec:Conclusion}

% In this work, we introduce full fine-tuning with SPT , compare different various SPT-based methods and integrate them to validate their effectiveness in low-resource transfer through CS ASR experiments, filling a gap in the comprehensive evaluation of SPT for ASR. Our proposed method achieves same  error reduction performance similar to FFT and LoRA approaches while effectively mitigating catastrophic forgetting, utilizing significantly fewer task-specific parameters than existing methods. This demonstrates the potential of SPT to enhance cross-lingual capabilities while maintaining computational and storage efficiency.

We introduce FFT with SPT, compare various SPT-based methods, and integrate them into SPT4ASR to assess their effectiveness in low-resource transfer through CS ASR, addressing the gap in comprehensive SPT evaluation for ASR. Our approach reduces errors and mitigates catastrophic forgetting, though it does not outperform FFT, while using fewer task-specific parameters, similar to LoRA. 
% This highlights SPT’s potential to improve cross-lingual capabilities while maintaining computational and storage efficiency. 
While LoRA remains more effective in terms of performance and parameter efficiency, soft prompt-based methods represent a viable alternative, offering a compelling trade-off between adaptability and resource usage, and warrant continued investigation in multilingual ASR.

% We also acknowledge that LoRA typically outperforms SPT4ASR in terms of parameter efficiency and performance. However, methods based on SPT can serve as a viable alternative to LoRA and are worth exploring. They provide a different approach to cross-lingual adaptation and, like LoRA, offer lower computational and storage costs compared to full fine-tuning. We believe that with ongoing technological advancements, soft prompt-based methods will continue to improve due to their parameter efficiency.

% We introduce full fine-tuning with SPT, compare multiple SPT-based methods, and integrate them into SPT4ASR to evaluate their effectiveness in low-resource transfer via code-switching ASR. While SPT4ASR does not outperform full fine-tuning (FFT), it significantly reduces errors and mitigates catastrophic forgetting, using fewer task-specific parameters similar to LoRA. Compared to LoRA, SPT-based methods offer a viable alternative for cross-lingual adaptation, maintaining lower computational and storage costs. Although LoRA typically achieves better performance and parameter efficiency, the flexibility and lightweight nature of soft prompt-based approaches make them a promising direction for future multilingual ASR systems.

\bibliographystyle{IEEEtran}
\bibliography{mybib}

\end{document}